\documentclass{article} 
\usepackage[preprint]{colm2026_conference}

\usepackage{microtype}
\usepackage{hyperref}
\usepackage{url}
\usepackage{booktabs}
\usepackage{subcaption}
\usepackage{algorithm}
\usepackage{algpseudocode}

\usepackage{graphicx}
\usepackage{amsmath} 
\usepackage{multirow}
\usepackage{tikz}
\usepackage{pgf-pie}
\usepackage{amssymb}
\usetikzlibrary{positioning,arrows.meta}
\usepackage[most]{tcolorbox}
\usepackage{xcolor}
\usepackage{wrapfig}
\definecolor{boxblue}{RGB}{176, 211, 255}
\definecolor{boxgrey}{RGB}{203, 203, 203}
\definecolor{boxgreen}{RGB}{182, 228, 106}
\definecolor{boxorange}{RGB}{249, 199, 179}
\definecolor{boxpurple}{RGB}{192, 231, 227}
\definecolor{Accent}{RGB}{20,60,200}

\newcounter{module}

\tcbset{
  agentstyle/.style={
    enhanced,
    colback=white,
    boxrule=1pt,
    arc=0pt, outer arc=0pt,
    left=6pt,right=6pt,top=8pt,bottom=6pt,
    borderline={0.8pt}{-1.8pt}{#1},
    attach boxed title to top left={xshift=6pt,yshift=-8pt},
    boxed title style={
      colback=#1, colframe=#1, boxrule=0pt,
      left=4pt,right=4pt,top=2pt,bottom=2pt
    },
    fonttitle=\small\bfseries,
    coltitle=black
  }
}

\newtcolorbox{topicseg}[2][]{agentstyle=boxblue,  colframe=boxblue,
  before upper={\refstepcounter{module}},
  title={#2}, #1}

\newtcolorbox{emo}[2][]{agentstyle=boxgrey,   colframe=boxgrey,
  before upper={\refstepcounter{module}},
  title={#2}, #1}

\newtcolorbox{speakersum}[2][]{agentstyle=boxpurple,   colframe=boxpurple,
  before upper={\refstepcounter{module}},
  title={#2}, #1}

\newtcolorbox{topicsum}[2][]{agentstyle=boxgreen, colframe=boxgreen,
  before upper={\refstepcounter{module}},
  title={#2}, #1}

\newtcolorbox{oversum}[2][]{agentstyle=boxorange, colframe=boxorange,
  before upper={\refstepcounter{module}},
  title={#2}, #1}

\definecolor{E1}{HTML}{4E79A7}
\definecolor{E2}{HTML}{F28E2B}
\definecolor{E3}{HTML}{E15759}
\definecolor{E4}{HTML}{76B7B2}
\definecolor{E5}{HTML}{59A14F}
\definecolor{E6}{HTML}{EDC948}
\definecolor{E7}{HTML}{B07AA1}
\definecolor{E8}{HTML}{FF9DA7}
\definecolor{E9}{HTML}{9C755F}
\definecolor{E10}{HTML}{BAB0AC}

\usepackage{lineno}

\definecolor{darkblue}{rgb}{0, 0, 0.5}
\hypersetup{colorlinks=true, citecolor=darkblue, linkcolor=darkblue, urlcolor=darkblue}

\title{Dialogue Summarization with Emotion Dynamics 
\newline Using Topic- and Participant-Centric Decomposition}

\author{Linyun Xiang, Mark Neerincx, Stephanie Tan\\
Department of Intelligent Systems, Delft University of Technology\\
\texttt{\{l.xiang,m.a.neerincx,s.tan-1\}@tudelft.nl}}

\begin{document}

\ifcolmsubmission
\linenumbers
\fi

\maketitle

\begin{abstract}

Existing text summarization research has focused much on monologic information (e.g., newspaper articles, reports) without accounting for the interaction between speakers or authors. In contrast, dialogues are a rich communication channel where multiple participants conduct back and forth exchanges to construct meaning. We propose a dialogue summarization framework that explicitly models both semantic and emotion dynamics using multimodal dialogue inputs, built on an adapted hierarchical Chain-of-Agents approach. We decompose dialogues from two perspectives: (1) topic segments based on the utterances of all participants, and (2) participant-specific utterance segments. These are used to generate corresponding summaries while incorporating automatically inferred emotions. Topic- and participant-level summaries are aggregated into a dialogue summary capturing semantic content and emotion trajectories. To evaluate beyond content accuracy, we introduce emotion trajectory metrics measuring how well summaries preserve emotional flow. Experiments with small language models on multimodal dialogue datasets show that our framework produces summaries with both semantic and emotion content. Further experiments on explicit emotion label availability highlight the efficacy of our proposed methodology and the opportunities in dialogue analysis using language models.


\end{abstract}

\section{Introduction}
Spoken interactions such as dialogues are frequent in collaborative settings, where information overload is common \citep{luong2005meetings, doi:10.1177/0963721418776307, cheng2024emotionllamamultimodalemotionrecognition}. Dialogue summarization has therefore emerged as a much-needed tool to help participants recall key points and enable absentees to stay informed \citep{Kumar2022MeetingSA}. However, dialogues are not only constructed utterance sequences containing semantic information. They are complex social interactions \citep{article} in which participants communicate not only through speech content but also engage through paralinguistic and nonverbal cues shaping the interaction. In particular, emotional signals are conveyed through multiple modalities, including tone, prosody, facial expressions, etc. \citep{ashforth1995emotion, annurev:/content/journals/10.1146/annurev-orgpsych-032516-113231, nikander2007emotions, barrett2007language}.

Despite this motivation, most existing dialogue summarization methods adopt a static, content-centric formulation. This is often operationalized by synthesizing summaries through extractive or abstractive information \citep{mishra-etal-2023-llm, gliwa-etal-2019-samsum}. Conversations are represented as fixed transcripts and several existing approaches have leveraged dialogue structure such as identifying salient topics or utterances to achieve summarization \citep{li-etal-2019-keep, zou2021topic, zhu-etal-2021-mediasum}. However, without explicitly accounting for conversation dynamics, these summaries do not contain descriptions of the underlying interaction processes, such as how emotional trajectories of the participants unfold throughout the dialogue. Beyond modeling emotional dynamics, the adoption of a perspectivist view on dialogues \citep{li2025grounded} and the aggregation of information based on participant-specific viewpoint remain largely underexplored in dialogue summarization \citep{feng2022surveydialoguesummarizationrecent}.

Large language models (LLMs) have demonstrated increasing capability in dialogue summarization \citep{mishra-etal-2023-llm, retkowski-waibel-2025-zero, tian-etal-2024-dialogue}, enabling both zero-shot and few-shot summarization of conversational content \citep{chhabra-etal-2024-revisiting, chen-etal-2023-unisumm}. Parallel to these developments in summarization, LLMs have been used to identify affective content \citep{10591796, 10888198}. Among these, multimodal large language models leveraging text, audio, and visual cues have been fine-tuned for emotion recognition \citep{sun2025dialoguemllm}. Furthermore, recent work on collaborative and role-specialized LLM agents highlights the potential of multi-agent systems to decompose complex language understanding tasks into modular, interpretable components \citep{2023arXiv230808155W, 2023arXiv230810848C}. Such approaches and their combinations suggest a promising pathway for using flexible multi-agent architectures to generate more nuanced summaries that integrate affective dynamics with semantic content.

In this paper, we study dialogue summarization with emotion dynamics, where the goal is to generate a summary that preserves both the semantic content and the emotional dynamics of a multi-party dialogue. To this end, we adapt a specific multi-agent framework: Chain-of-Agents (CoA) \citep{zhang2024chain} to coordinate multiple LLM-based agents for layered information aggregation. Our framework combines separate LLM and MLLM components for textual and multimodal inputs, using a hierarchical organization across utterance-, topic segment-, participant-, and dialogue-level processing. 

Our contributions are threefold: \textbf{(1)} a hierarchical CoA framework for multimodal dialogue summarization that captures semantic and emotional dynamics, with novel metrics for emotion trajectory preservation; \textbf{(2)} an analysis of how different dialogue decompositions (topic and participant views) affect summary quality and emotional trajectory preservation; and \textbf{(3)} an investigation of how explicit emotion labels in instructions influence summary quality and trajectory fidelity, which is particularly relevant in a human centered setting where emotion information may be restricted due to privacy and user preferences.


\section{Related Work}

\paragraph{Content-centric dialogue summarization}

Early work on dialogue summarization focused on identifying and condensing salient informational content from spoken interactions, treating conversations as collections of utterances analogous to multi-document or spoken-document summarization \citep{RIEDHAMMER2010801}. Classical approaches relied on supervised learning and discourse-informed features, such as topic modelling \citep{WU201712}, speaker turns \citep{BOKAEI_SAMETI_LIU_2016}, and dialogue acts \citep{SUN2024103635}.
Extractive techniques based on ranking \citep{galley-2006-skip} or graph-based representations \citep{garg2009clusterrank, 2024arXiv240511055R} were widely adopted. Neural approaches enabled abstractive dialogue summarization beyond sentence extraction. Models based on BART \citep{47751}, T5 \citep{raffel2023exploringlimitstransferlearning}, and domain-adapted architectures \citep{zhong-etal-2021-qmsum, zhu2020hierarchical} demonstrated improvements in fluency and coherence. More recently, pretrained LLMs have further lowered the barrier to producing high-quality abstractive summaries, making summarization more scalable and accessible \citep{2024arXiv240611289Z, 10.1016/j.neucom.2024.128255}.

Despite these advances, content-centric summarization systems still primarily emphasize semantic content, often overlooking emotional aspects of conversation \citep{zhou2024psentscoreevaluatingsentimentpolarity}. As a result, they fail to account for the social nuances and interactional dynamics that shape how discussions develop. Moreover, unlike single-author documents, dialogues involve multiple participants whose perspectives and interactions shape the progression of discussions, making summarization inherently more complex \citep{feng2022surveydialoguesummarizationrecent}. As noted by \citet{liang-etal-2023-enhancing, zhong-etal-2021-qmsum}, both participants’ viewpoints and discussion topics evolve throughout the dialogue stream, suggesting that conversational structure extends beyond topical content alone.

However, existing summarization approaches largely rely on a single structural perspective. They typically focus on topic-level representations and rarely model how conversational dynamics and participant viewpoints jointly influence the development of discussions.

\paragraph{Interaction modeling and emotion dynamics in dialogue}

Parallel to advances in dialogue summarization, some studies have focused on modeling conversational interactions. Research in dialogue analysis and social signal processing characterizes conversation as a dynamic social process shaped by turn-taking patterns \citep {Sacks1974a}, speaker intentions \citep{bunt-etal-2010-towards}, and evolving emotional states \citep{poria-etal-2019-meld}. Advances in LLMs have further enabled scalable modeling of such interactional phenomena. Existing studies address interaction-focused tasks including conversation forecasting \citep{li2025onlinemultimodalsocialinteraction}, speaker intent modeling \citep{Yin2024ECLMEL}, sentiment analysis \citep{luo2024panosentpanopticsextupleextraction}, and multi-speaker attention alignment \citep{2025arXiv251117952O}, emphasizing dialogue properties that extend beyond topical content.

Emotion recognition in dialogue represents a notable line of work. Approaches such as DialogueRNN \citep{majumder2019dialoguernnattentivernnemotion} and DialogueGCN \citep{ghosal-etal-2019-dialoguegcn} explicitly model speaker-dependent emotional transitions throughout dialogues. Recent work has also explored conversational dynamics from a purely textual perspective \citep{hua-etal-2024-get}. However, because these approaches rely exclusively on text, they are inherently limited when applied to face-to-face interactions, where rich signals such as speech prosody, facial expressions, and body gestures play a crucial role \citep{VINCIARELLI20091743}. In such multimodal settings, emotion cues inferred solely from transcripts provide an incomplete view of the situated interaction \citep{poria2017review}. More recent work on MLLMs has extended emotion modeling to multimodal settings by incorporating audio-visual cues \citep{yang2025omniemotionextendingvideomllm, sun2025dialoguemllm}, enabling the capture of affective and interactional signals unavailable from text alone \citep{poria2017review}.

Despite progress in dialogue summarization and emotion modeling, these areas have largely evolved separately. Multimodal emotion work focuses on affective cues, while summarization remains content centric. To the best of our knowledge, existing approaches have not systematically integrated multimodal emotion perception into dialogue summarization, leaving open how emotion dynamics can be reflected in generated summaries.

\section{Approach}

\begin{figure*}[h]
    \centering
    \includegraphics[width=\textwidth]{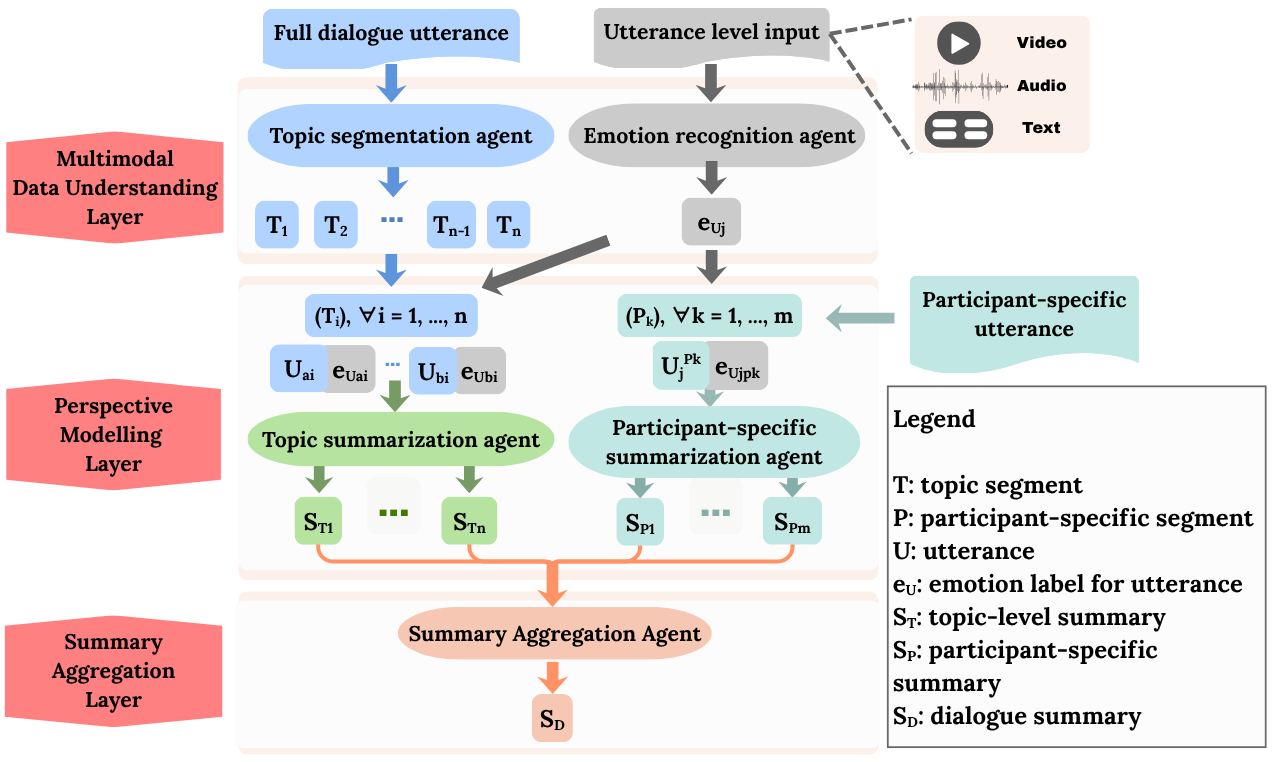}
    \caption{Overview of the proposed dialogue summarization framework with Multimodal Data understanding, Perspective Modeling, and Summary Aggregation Layer.}
    \label{fig:approach}
\end{figure*}

Dialogue summarization with emotion dynamics requires capturing not only the semantic content of the conversation but also the emotional states expressed by participants as the dialogue unfolds.To address this challenge, we caste the task as a Chain-of-Agents problem \citep{zhang2024chain}. Instead of using a single monolithic model to jointly perform topic segmentation, infer emotional dynamics, and generate a global summary, the task is decomposed such that specialized agents perform constrained subtasks. Results obtained from these structured subtasks are passed to other agents downstream. 
This design benefits from having multiple specialized small language models that enable us to parallelize the processing of the semantic and emotion-related subtasks, while accounting for both topic- and participant-centric perspectives of the dialogue. This also allows for the final summary to reflect not only what was discussed but also how participants are emotionally engaged throughout the dialogue. The overall framework is illustrated in Figure~\ref{fig:approach}. This framework consists of three layers: multimodal data understanding, perspective modelling and summary aggregation, which enables progressive abstraction.



The \textbf{\textit{Multimodal Data Understanding Layer}} processes multimodal inputs such as text transcript, and audio and video of the participants. It comprises two agents.  First, a \textbf{Topic Segmentation Agent} denoted as $f_T$, partitions a dialogue comprised of utterances $D = \{U_j\}^{L}_{j=1}$ into $n$ coherent topical segments, following topic segmentation prompt instruction $I_\text{topic segmentation}$. The agent returns the utterance boundaries for each topic, 
\begin{equation}
f_T(D, I_\text{topic segmentation}) = \{T_i:(a_i, b_i)\}_{i=1}^n ,
\end{equation}
\noindent where $(a_i, b_i)$ are the start and end utterance indices of topic $T_i$. Second, an \textbf{Emotion Recognition Agent} $f_E$, given emotion recognition prompt instruction $I_\text{emotion recognition}$, processes utterance-level emotions with multimodal inputs of text, audio, and video of the $j$-th utterance of a participant, denoted by $(U_j, A_j, V_j)$ respectively. The agent $f_E$ generates an emotion prediction using multimodal information associated with each participant utterance, defined as
\begin{equation}
f_E(U_j, A_j, V_j, I_\text{emotion recognition}) = e_{U_j} ,
\end{equation}
where 
$e_{U_j} \in \mathcal{E} $ is the predicted emotion label for utterance $j$, and $\mathcal{E}$ =\{neutral, happiness, sadness, surprise, excited, frustration, anger\}, inspired from the emotion categories used in affective research {\citep{Ekman01051992, picard2000affective, busso2008iemocap}}.  As a result, the Multimodal Data Understanding Layer produces explicit topic segments consisting utterances and emotion labels per utterance for the downstream Perspective Modeling layer.

The \textit{\textbf{Perspective Modeling Layer}} operationalizes generating summaries from two structured viewpoints of the dialogue: a topic-centric perspective and a participant-centric perspective. 
The topic-centric perspective is modeled using a \textbf{Topic Summarization Agent}, denoted by $f_{S^{\text{topic}}}$. It utilizes topics identified by the aforementioned Topic Segmentation Agent to produce topic-wise summaries that capture emotional dynamics within each topic. Concretely, for a given topic $T_i$ and its topic boundaries $(a_i,b_i)$, we collect all the utterances and their respective emotion predictions, denoted as $\{(U_j, e_{U_j})\}_{j=a_i}^{b_i}$. The summarization agent then generates 
a topic summary using the prompt instruction $I_{\text{topic summarization}}$:
\begin{equation}
f_{S^{\text{topic}}} (\{(U_j, e_{U_j})\}_{j=a_i}^{b_i}, I_{\text{topic summarization}}) = S_{T_i}, \forall i = 1, \dots, n,
\end{equation}
where $S_{T_i}$ denotes the summary of topic $T_i$. Concurrently, a \textbf{Participant-specific Summarization Agent} denoted as $f_{S^{\text{participant}}}$, 
generates participant summaries by using utterances and corresponding emotion labels of a specific participant. Let  $\mathcal{P} = \{P_1, P_2, \dots, P_m\}$ denote the participant set where $m$ is the total number of participants in the dialogue. For each participant $P_k \in \mathcal{P}$, we define 
$\{(U_j^{P_k}, e_{{U_j}^{P_k}})\}$ as a set of utterances and their emotion predictions. Along with prompt instruction $I_{\text{participant summarization}}$, we obtain $S_{P_k}$ denoting  the summary associated with participant $P_k$,
\begin{equation}
f_{S^{\text{participant}}}(\{(U_j^{P_k}, e_{{U_j}^{P_k}})\}, I_{\text{participant summarization}}) = S_{P_k}, \forall k = 1, \dots, m.
\end{equation}

Finally, the \textit{\textbf{Summary Aggregation Layer}} includes a \textbf{Summary Aggregation Agent} $f_{S^{\text{dialogue}}}$, which integrates the topic-level summaries $\{S_{T_i}\}_{i=1}^n$ and participant-specific summaries $\{S_{P_k}\}_{k=1}^m$ into a coherent, global dialogue summary. This final summary aims to preserve both key semantic content and the evolution of emotional dynamics of the dialogue. Using a prompt instruction $I_{\text{dialogue summarization}}$, the overall summary $S_D$ is defined as
\begin{equation}
f_{S^{\text{dialogue}}}\!\left(\{S_{T_i}\}_{i=1}^{n}, \{S_{P_k}\}_{k=1}^{m}, I_{{\text{dialogue summarization}}} \right) = S_{D}.
\end{equation}
The overall framework can thus be viewed as a structured multi-layer pipeline in which $f_T$ produces topic boundaries, $f_E$ predicts utterance-level emotions, $f_{S^{\text{topic}}}$ generates topic-level summaries, $f_{S^{\text{participant}}}$ generates participant-specific summaries, and $f_{S^{\text{dialogue}}}$ aggregates both into a dialogue summary, represented by 
$f_{S^{\text{dialogue}}} \circ f_{S^{\text{participant}}} \circ f_{S^{\text{topic}}} \circ f_E \circ f_T (D) = S_D$. See illustrative examples of outputs $T_i$, $E_{U_j}$, $S_{T_i}$, $S_{P_k}$ and $S_D$ in Appendix~\ref{app:agent_example}. 

\section{Datasets and Implementation}
To validate our approach, we consider multimodal dialogue datasets with audio and video modality coverage. Among existing datasets of multimodal dialogues \citep{icsi, hu-etal-2023-meetingbank, poria-etal-2019-meld, carletta2005ami, busso2008iemocap}, we select the AMI and IEMOCAP datasets. AMI is a meeting dataset with approximately 140 4-person meetings. IEMOCAP captures multimodal dialogues of dyadic interactions, which consist of 150 scripted dialogues and their spontaneous enactment. From our purpose of multimodal emotion perception, we require each dialogue session to contain personalized audio and video streams for all participants. After this filtering, the AMI dataset contains 26 dialogues and IEMOCAP contains 150 dialogues with full multimodal information. 

We conduct statistical analyses covering average duration, word count, and token count at both the utterance and topic levels (Table \ref{tab:dataset-distribution}). These statistics characterize structurally distinct dialogue settings: AMI consists of long, information-dense meeting segments dominated by neutral affect, whereas IEMOCAP comprises shorter, more emotionally expressive exchanges with a broader distribution of emotions. Making these properties explicit supports transparent interpretation of results and discourages overgeneralization across datasets with differing interactional characteristics, as we explain in Section~\ref{sec:experiment} (RQ3).

Importantly, AMI and IEMOCAP differ in the availability of ground-truth annotations. As our proposed task of dialogue summarization with emotion trajectories is novel, none of these datasets have matching ground truth summaries. Topic boundaries are annotated in AMI but not in IEMOCAP; and emotions are annotated for IEMOCAP but not for AMI. Therefore, we use off-the-shelf existing models to obtain topic segments (LLaMA 3.1 8B) and emotions (Emotion-LLaMA)\citep{cheng2024emotionllamamultimodalemotionrecognition}, for IEMOCAP and AMI, respectively. In this context, the performance of both Topic Segmentation and Emotion Recognition agents is reported in Appendix~\ref{sec:appendix_agents}. While this setup affects our Multimodal Data Understanding Layer, its does not affect our main investigation of generating summaries with emotion trajectories with different dialogue decomposition views (topic vs. participants), and with emotion label availability in the instructions. To avoid confounding influences in our main investigations, we use ground-truth topic boundaries and emotions for the downstream pipeline whenever available. See the results of using a fully automated pipeline in Appendix~\ref{tab:full_pipeline}. 

The remaining components of the framework (i.e., Topic Summarization Agent, Participant-Specific Summarization agent, and Summary Aggregation Agent) are built on the LLaMA family \citep{touvron2023llamaopenefficientfoundation}, chosen for strong performance in text summarization and its ability to generate coherent, well-structured outputs \citep{SAXENA2025532}. Prior studies show that LLaMA performs competitively with other open-source models and serves as an effective alternative to proprietary systems such as GPT-4 for complex summarization tasks \citep{aly2025evaluationlargelanguagemodels,10826032}. We use LLaMA 3.2 3B and LLaMA 3.1 8B to demonstrate our methodological approach; benchmarking with other models is left for future studies. Evaluation is conducted using LLaMA 3.1 8B to extract speaker-wise emotion trajectories from generated summaries. The model is prompted to classify emotions expressed by each participant, enabling comparison with ground-truth emotion sequences. All models are implemented using default settings, except for the context window of the LLaMA-based models, which is extended to 32\,467 tokens to accommodate long dialogue inputs.

\section{Evaluation metrics}
\paragraph{Content-level summary evaluation}
As no ground-truth annotations of summary with embedded emotion trajectories are available for this task, to assess summary informativeness with respect to content, we use the $\text{BLANC}_{\text{help}}$ metric \citep{vasilyev2020blanchumanfreequalityestimation}, which is a reference-free metric that evaluates how well a summary (i.e., summary $S_D$ in our case) helps a pretrained language model reconstruct masked tokens in the source document (i.e., dialogue $D$ in our case). Let $\text{Acc}_{\text{base}}$ denote the masked token prediction accuracy without access to the summary, and $\text{Acc}_{\text{help}}$ denote the accuracy when the summary is provided. The BLANC-help score is defined as the improvement in prediction accuracy: $\text{BLANC}_{\text{help}}
=
\text{Acc}_{\text{help}} - \text{Acc}_{\text{base}}.$
Higher scores indicate higher summary quality. While BLANC scores could range from -1 to 1, typical range is 0 - 0.3 \citep{vasilyev2020fill}.  


\paragraph{Emotion Trajectory Evaluation} To evaluate predicted emotion trajectories, we compare model-predicted emotion sequences from the generated summaries with ground-truth emotion sequences at the participant level. Predicted trajectories are obtained using LLaMA 3.1 8B by prompting the model to extract emotion sequences from summaries, with emotion being a class within aforementioned $\mathcal{E}$.
Let the raw ground-truth and predicted emotion trajectories be $\mathbf{e}^{\text{gt}}$ and $\mathbf{e}^{\text{pred}}$. To focus on emotion transitions rather than repeated consecutive states, we construct compressed trajectories using an algorithm~\ref{alg:compress_emotion}, yielding $\hat{\mathbf{e}}^{\text{gt}}$ and $\hat{\mathbf{e}}^{\text{pred}}$. All similarity metrics are computed over the compressed trajectories.

As no standard metrics exist for this task, we adopt a set of complementary measures that capture global sequence alignment (Levenshtein), local temporal patterns ($n$-grams), transition coverage (Jaccard), and transition distributional similarity (cosine).
\textbf{Normalized Levenshtein similarity} \citep{1966SPhD...10..707L} is defined as $1 - \frac{\text{LevDistance}(\hat{\mathbf{e}}^{\text{pred}}, \hat{\mathbf{e}}^{\text{gt}})}{\max(|\hat{\mathbf{e}}^{\text{pred}}|, |\hat{\mathbf{e}}^{\text{gt}}|)}$
where $|\cdot| $ represents the length of the sequence. It measures global sequence alignment capturing ordering, substitutions, and length differences. 
We evaluate local and transition-level alignment using subsequence-based metrics derived from compressed emotion trajectories. Let 
\(\mathcal{T}_x(\hat{\mathbf{e}})\) denote the set of all contiguous subsequences of length \(x\) extracted from \(\hat{\mathbf{e}}\). In particular, \(\mathcal{T}_1\) corresponds to unigrams, \(\mathcal{T}_n\) to $n$-grams, and \(\mathcal{T}_2\) to transition pairs.The \textbf{$n$-gram overlap} \citep{brown-etal-1992-class} is defined as \(\frac{|\mathcal{T}_n(\hat{\mathbf{e}}^{\text{pred}}) \cap \mathcal{T}_n(\hat{\mathbf{e}}^{\text{gt}})|}{|\mathcal{T}_n(\hat{\mathbf{e}}^{\text{gt}})|}\), for \(n \in \{1,2,3\}\), measuring how well local emotion patterns are preserved.
\textbf{Jaccard similarity} \citep{jaccard1901etude} is defined over transition sets (\(x=2\)) as 
\(\frac{|\mathcal{T}_2(\hat{\mathbf{e}}^{\text{pred}}) \cap \mathcal{T}_2(\hat{\mathbf{e}}^{\text{gt}})|}{|\mathcal{T}_2(\hat{\mathbf{e}}^{\text{pred}}) \cup \mathcal{T}_2(\hat{\mathbf{e}}^{\text{gt}})|}\).
\textbf{Cosine similarity} \citep{10.5555/576628} is defined as 
\((\mathbf{v}^{\text{pred}} \cdot \mathbf{v}^{\text{gt}}) \, / \, (\|\mathbf{v}^{\text{pred}}\| \|\mathbf{v}^{\text{gt}}\|)\), 
where \(\mathbf{v}\) denotes the frequency vector over all possible transition pairs from \(\mathcal{T}_2(\hat{\mathbf{e}})\), and \(\|\cdot\|\) denotes the Euclidean (L2) norm, \(\|\mathbf{v}\| = \sqrt{\sum_i v_i^2}\). Higher scores on all metrics indicate stronger alignment between predicted and ground-truth emotion trajectories. Results are averaged across participants and dialogues.

\section{Experiment}
\label{sec:experiment}

We assess the quality of the generated dialogue summaries and whether or not they preserve both semantic content and emotion trajectories from the dialogue in the AMI and IEMOCAP datasets, focusing three research questions (RQ). 
\noindent

\textbf{RQ1:} How do different dialogue decomposition strategies affect the quality of the generated global summaries and their ability to preserve emotional trajectories?

We consider two decomposition strategies to generate summaries: topic-level segmentation and participant-level segmentation. This is illustrated as Perspective Modeling Layer in our approach. Topic-level summaries capture the chronological progression of discussion segments and reflect how the conversation evolves across topics. In contrast, participant-specific summaries aggregate all utterances belonging to each participant, providing a speaker-centric view that highlights individual perspectives and emotional developments throughout the dialogue. To evaluate the effect of these segmentation strategies on the final summary, we compare three aggregation settings as inputs to the final Summary Aggregation Agent: Topic-only, Participant-only, and Combined aggregation.

\begin{table*}[t]
\centering
\small
\resizebox{\textwidth}{!}{
\begin{tabular}{l l ccccc ccccc}
\toprule

\multirow{3}{*}{\textbf{Model}} &
\multirow{3}{*}{\textbf{Setup}} &
\multicolumn{5}{c}{\textbf{AMI}} &
\multicolumn{5}{c}{\textbf{IEMOCAP}} \\

\cmidrule(lr){3-7} \cmidrule(lr){8-12}

& &
\multicolumn{1}{c}{\textbf{Content}} &
\multicolumn{4}{c}{\textbf{Emotion}} &
\multicolumn{1}{c}{\textbf{Content}} &
\multicolumn{4}{c}{\textbf{Emotion}} \\

\cmidrule(lr){3-3} \cmidrule(lr){4-7} \cmidrule(lr){8-8} \cmidrule(lr){9-12}

& &
\textbf{BL} & \textbf{LEV} & \textbf{NGR} & \textbf{JAC} & \textbf{COS} &
\textbf{BL} & \textbf{LEV} & \textbf{NGR} & \textbf{JAC} & \textbf{COS} \\

\midrule

\multirow{2}{*}{LLaMA 3.2 3B}
& Topic       & 0.122 & 0.303 & 0.409 & 0.285 & 0.296 & 0.097 & 0.460 & 0.503 & 0.357 & 0.403 \\
& Participant & 0.057 & 0.168 & 0.335 & 0.215 & 0.293 & 0.089 & 0.321 & 0.413 & 0.221 & 0.334 \\

\midrule

\multirow{2}{*}{LLaMA 3.1 8B}
& Topic       & 0.102 & 0.356 & 0.507 & 0.361 & 0.375 & 0.106 & 0.489 & 0.533 & 0.389 & 0.432 \\
& Participant & 0.062 & 0.167 & 0.503 & 0.306 & 0.419 & 0.091 & 0.350 & 0.547 & 0.268 & 0.400 \\

\bottomrule
\end{tabular}
}
\caption{Summarization results at the Perspective Modeling Layer on AMI and IEMOCAP.}
\label{tab:rq1_intermediate}
\end{table*}

\begin{table*}[t]
\centering
\small
\resizebox{\textwidth}{!}{
\begin{tabular}{l l ccccc ccccc}
\toprule

\multirow{3}{*}{\textbf{Model}} &
\multirow{3}{*}{\textbf{Setup}} &
\multicolumn{5}{c}{\textbf{AMI}} &
\multicolumn{5}{c}{\textbf{IEMOCAP}} \\

\cmidrule(lr){3-7} \cmidrule(lr){8-12}

& &
\multicolumn{1}{c}{\textbf{Content}} &
\multicolumn{4}{c}{\textbf{Emotion}} &
\multicolumn{1}{c}{\textbf{Content}} &
\multicolumn{4}{c}{\textbf{Emotion}} \\

\cmidrule(lr){3-3} \cmidrule(lr){4-7} \cmidrule(lr){8-8} \cmidrule(lr){9-12}

& &
\textbf{BL} & \textbf{LEV} & \textbf{NGR} & \textbf{JAC} & \textbf{COS} &
\textbf{BL} & \textbf{LEV} & \textbf{NGR} & \textbf{JAC} & \textbf{COS} \\

\midrule

\multirow{3}{*}{LLaMA 3.2 3B}
& Topic       & 0.013 & 0.062 & 0.469 & 0.207 & 0.240 & 0.054 & 0.247 & 0.407 & 0.160 & 0.247 \\
& Participant & 0.045 & 0.062 & 0.386 & 0.147 & 0.164 & 0.060 & 0.252 & 0.473 & \textbf{0.193} & 0.274 \\
& Combined    & \textbf{0.058} & \textbf{0.094} & \textbf{0.503} & \textbf{0.222} & \textbf{0.257} & \textbf{0.070} & \textbf{0.265} & \textbf{0.483} & 0.192 & \textbf{0.290} \\

\midrule

\multirow{3}{*}{LLaMA 3.1 8B}
& Topic       & 0.046 & 0.099 & 0.507 & 0.236 & 0.278 & 0.066 & 0.286 & 0.447 & 0.208 & 0.310 \\
& Participant & 0.050 & 0.079 & 0.485 & 0.203 & 0.240 & 0.054 & 0.277 & 0.429 & 0.204 & 0.300 \\
& Combined    & \textbf{0.065} & \textbf{0.106} & \textbf{0.540} & \textbf{0.250} & \textbf{0.295} & \textbf{0.081} & \textbf{0.286} & \textbf{0.468} & \textbf{0.216} & \textbf{0.313} \\

\bottomrule
\end{tabular}
}
\caption{Summarization results at the Summary Aggregation Layer on AMI and IEMOCAP. These are the final dialogue summary outcome of our framework. }
\label{tab:rq1_final}
\end{table*}

Table~\ref{tab:rq1_intermediate} presents the summarization results at the Perspective Modeling Layer, where topic-level and participant-specific summaries are evaluated. Topic-based summaries consistently outperform participant-based summaries across both models and datasets, indicating that topic-level structuring provides more informative intermediate results. This trend is particularly evident in the emotion-related metrics, whereas differences in BLANC (BL) scores are comparatively smaller, suggesting that content preservation is less sensitive to the decomposition strategy at this stage.

However, at the final Summary Aggregation Layer (Table~\ref{tab:rq1_final}), combining topic and participant information generally leads to improved performance across most metrics and datasets. The Combined setup often achieves the highest or near-highest scores. This suggests that topic and participant perspectives may capture different aspects of the dialogue, and their integration can be beneficial for generating higher-quality global summaries. Notably, BLANC again shows relatively limited variation across setups, while the emotion-related metrics exhibit more pronounced differences, highlighting that the benefits of combination are more evident in modeling emotional consistency than in content overlap.

Also interestingly in Table~\ref{tab:rq1_final}, while topic-based summaries dominate at the Perceptive Modeling Layer, using participant-based summaries become more competitive for generating overall dialogue summaries, particularly for emotion-related metrics (e.g., higher COS scores on IEMOCAP). This indicates that participant-level information may better preserve emotional trajectories, which may be refined during the final summary aggregation step.

\noindent
\textbf{RQ2:} How does emotion information in instructions affect summary quality and alignment with dialogue emotion dynamics?

In many real-world scenarios, utterance-level emotion labels may not always be available. Emotion labels can be missing due to system limitations (e.g., feasibility of obtaining these in real-time) or privacy constraints due to the sensitive nature of emotions in dialogues. Therefore, it is important to understand how dialogue summarization systems perform under the presence or absence of emotion-related input instructions.

To investigate this question, we examine two factors: the availability of utterance-level emotion labels ($E$) and the presence of explicit emotion trajectory-preservation instructions ($T$). The $E^+$ and $E^-$ conditions test whether providing explicit emotion labels improves summarization performance compared to settings where such information is unavailable. The $T^+$ and $T^-$ conditions examine whether explicitly instructing the model to preserve emotional trajectories influences the generated summaries compared to cases where no such guidance is provided. This results in a $2 \times 2$ experimental design that varies the presence of emotion annotations and trajectory instructions, as shown in Table~\ref{tab:rq2_setup}.
\begin{wraptable}{r}{0.5\textwidth}
\centering
\small
\begin{tabular}{p{1.0cm} p{1.25cm} p{3.8cm}}
\toprule
\textbf{Emotion Labels} & \textbf{Trajectory Instruction} & \textbf{Description} \\
\midrule
E-- & T-- & Transcript only with speaker labels. \\
E-- & T+  & Transcript only; prompt instructs the model to preserve emotional dynamics. \\
E+  & T-- & Transcript with utterance-level emotion labels; no trajectory instruction. \\
E+  & T+  & Transcript with utterance-level emotion labels and instruction to preserve emotion trajectories. \\
\bottomrule
\end{tabular}
\caption{Experimental configurations for RQ2.}
\label{tab:rq2_setup}
\end{wraptable}

 Table~\ref{tab:emotion_setup_results} evaluates the role of explicit emotion labels and emotion trajectory instruction in the combined topic and participant summaries setting. Incorporating both components (E+/T+) leads to the best overall performance, particularly for emotion-related metrics. For instance, in the LLAMA 3.1 8B model, E+/T+ achieves the highest scores on AMI (NGR: 0.540, COS: 0.295) and strong performance on IEMOCAP. In contrast, BLANC scores vary less across configurations. 
 This finding extends the earlier observation, indicating that content preservation is relatively insensitive to the availability of emotion information.

Other settings yield mixed results. E-/T+ and E+/T- improve summarization performance compared to the baseline E-/T-. Comparison between E-/T+ and E-/T- shows that even when the emotion labels are unavailable as inputs, our trajectory instructions still preserve the emotion trajectories better in the final summary.  Overall, these results suggest that utterance-level emotion labels and trajectory-preservation instructions are essential: combining both leads to more robust improvements, particularly for capturing emotional coherence in generated summaries.

\begin{table*}[h]
\centering
\small
\resizebox{\textwidth}{!}{
\begin{tabular}{l l ccccc ccccc}
\toprule

\multirow{3}{*}{\textbf{Model}} &
\multirow{3}{*}{\textbf{Setup}} &
\multicolumn{5}{c}{\textbf{AMI}} &
\multicolumn{5}{c}{\textbf{IEMOCAP}} \\

\cmidrule(lr){3-7} \cmidrule(lr){8-12}

& &
\multicolumn{1}{c}{\textbf{Content}} &
\multicolumn{4}{c}{\textbf{Emotion}} &
\multicolumn{1}{c}{\textbf{Content}} &
\multicolumn{4}{c}{\textbf{Emotion}} \\

\cmidrule(lr){3-3} \cmidrule(lr){4-7} \cmidrule(lr){8-8} \cmidrule(lr){9-12}

& &
\textbf{BL} & \textbf{LEV} & \textbf{NGR} & \textbf{JAC} & \textbf{COS} &
\textbf{BL} & \textbf{LEV} & \textbf{NGR} & \textbf{JAC} & \textbf{COS} \\

\midrule

\multirow{4}{*}{LLaMA 3.2 3B}
& {E-/T-}  & 0.058 & 0.065 & 0.325 & 0.113 & 0.155 & \textbf{0.073} & 0.192 & 0.383 & 0.123 & 0.184 \\
\cmidrule(lr){2-12}
& {E-/T+}  & 0.058 & 0.081 & 0.385 & 0.157 & 0.174 & 0.072 & 0.225 & 0.351 & 0.130 & 0.189 \\
\cmidrule(lr){2-12}
& {E+/T-}  & \textbf{0.059} & 0.077 & 0.381 & 0.143 & 0.191 & 0.071 & 0.256 & 0.469 & 0.182 & 0.277 \\
\cmidrule(lr){2-12}
& {E+/T+}  & 0.058 & \textbf{0.094} & \textbf{0.503} & \textbf{0.222} & \textbf{0.257} & 0.070 & \textbf{0.265} & \textbf{0.483} & \textbf{0.192} & \textbf{0.290} \\

\midrule

\multirow{4}{*}{LLaMA 3.1 8B}
& {E-/T-}  & \textbf{0.072} & 0.065 & 0.377 & 0.148 & 0.167 & \textbf{0.089} & 0.214 & 0.382 & 0.131 & 0.203 \\
\cmidrule(lr){2-12}
& {E-/T+}  & 0.063 & 0.082 & 0.437 & 0.196 & 0.223 & 0.082 & 0.225 & 0.353 & 0.131 & 0.197 \\
\cmidrule(lr){2-12}
& {E+/T-}  & 0.067 & 0.077 & 0.365 & 0.138 & 0.183 & 0.080 & 0.263 & \textbf{0.470} & 0.192 & 0.287 \\
\cmidrule(lr){2-12}
& {E+/T+}  & 0.065 & \textbf{0.106} & \textbf{0.540} & \textbf{0.250} & \textbf{0.295} & 0.081 & \textbf{0.286} & 0.468 & \textbf{0.216} & \textbf{0.313} \\

\bottomrule
\end{tabular}
}
\caption{
Effect of emotion-related information on dialogue summarization performance.
}
\label{tab:emotion_setup_results}
\end{table*}




\begin{figure}[t]
    \centering
    \includegraphics[width=\linewidth]{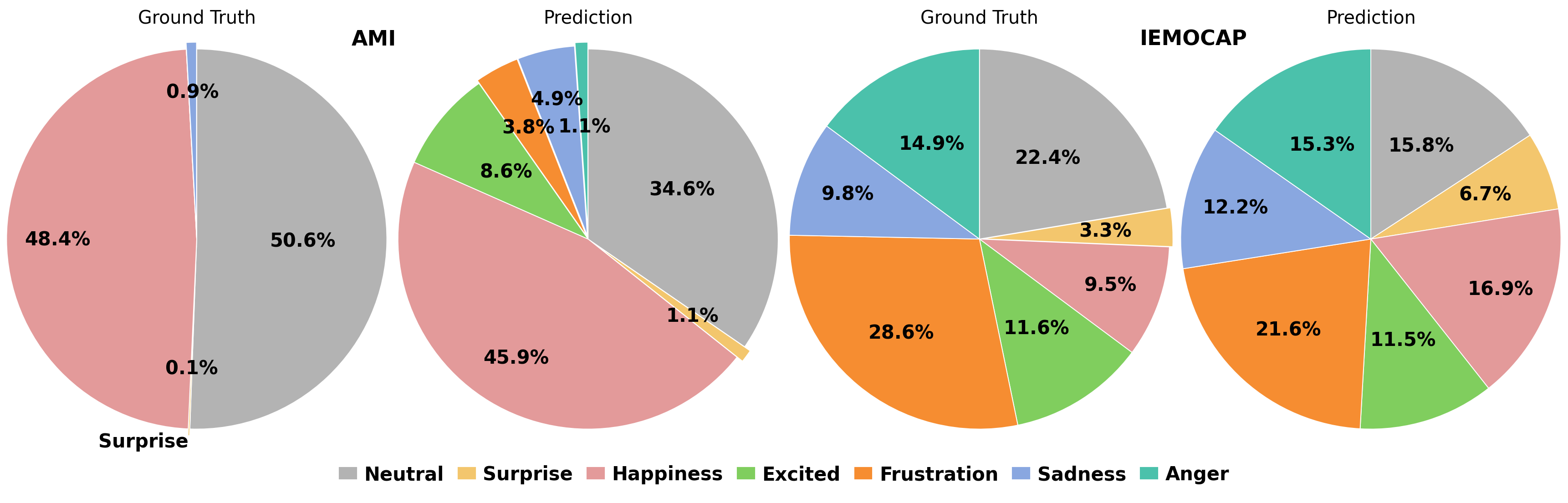}
    
    \caption{Percentage of emotions classes in ground truth and predicted emotion trajectories.}
    \label{fig:emotion_distribution}
\end{figure}

\noindent
\textbf{RQ3:} How does dialogue summarization performance differ with respect to emotion diversity represented in emotion trajectories?

The two datasets, AMI and IEMOCAP,  differ fundamentally in their interaction settings. \textsc{IEMOCAP} consists of dyadic, emotionally expressive conversations,
with balanced distribution across multiple emotion categories. In contrast, \textsc{AMI} captures four-person multi-party meetings.
Figure \ref{fig:emotion_distribution} illustrates the distribution of emotion classes across all ground truth and prediction trajectories. AMI has a majority of "Neutral'' and ``Happiness'' in ground truth trajectories. This imbalance may have stemmed from the inherently task-oriented nature of meetings or resulted from the performance gap in the current state-of-the-art emotion recognition MLLM (see Figure~\ref{fig:emotion_analysis} for emotion classification results on IEMOCAP).



These dataset differences are also reflected in both the frequency and diversity of emotion transitions in the trajectories. AMI exhibits a higher number of ground-truth emotion transitions per speaker (19.9 on average). Transitions between neutral and happiness occur frequently, averaging 19.5 bidirectional switches per speaker. In contrast, IEMOCAP features fewer transitions per speaker (8.1) with greater diversity across emotion categories. To analyze the relationship between model performance and emotion transition variance, we examine AMI ground-truth trajectories with transition counts between 7 and 9, thereby controlling for differences in transition frequency with IEMOCAP. Under comparable transition lengths, we compare summaries generated by LLaMA 3.1 8B model under the Combined (topic and participant) setting on AMI and IEMOCAP. Out of the emotion-related metrics, the LEV score on IEMOCAP is higher than that of AMI ($\Delta = 0.111$).  The NGR, JAC, and COS scores are higher for AMI than for IEMOCAP ($\Delta = 0.301, 0.207, 0.133$, respectively). Recall that LEV evaluates the overall trajectory quality while NGR, JAC, and COS evaluates local patterns and transition preservation in the sequence. Therefore, this suggests that the model better preserves local transition structure when emotion trajectories exhibit lower variance, as in AMI. When trajectories have higher variance, with more diverse emotion classes as in IEMOCAP, the model better captures the overall trajectory.




{\section{Discussions and Conclusions}
To conclude, we present a dialogue summarization framework that explicitly models emotion dynamics alongside semantic content, enabling summaries to capture both what is discussed and how interactions evolve emotionally. Through experiments with small language models, we demonstrate that emotion-aware summarization benefits from combining both dialogue decomposition perspectives (topic and participant) and the inclusion of explicit emotion input in instruction. 


While we analyze different design choices, the performance and utility of the framework remain to be characterized. In our experiments, the LLaMA 3.1 8B model consistently outperforms the LLaMA 3.2 3B model on both content and emotion metrics, with larger gains for emotion measures. However, we rely on pretrained small language models, and further improvements may be possible through fine-tuning. Our framework assumes access to ground truth speaker labels and segmented utterances. In practice, this requires a speaker diarization component in the audio pipeline and is susceptible to noisy environments. The MLLM emotion recognition module also requires individual frontal-facing video and audio, which may not always be high quality or available. Finally, our hierarchical design, while employing specialized agent models for subtasks, may introduce information loss during intermediate aggregation stages. Compressing dialogues into structured summaries based on topics can overlook long-range dependencies and subtle speaker dynamics. Further experimentation can be done with alternative multi-agent coordination strategies. Despite this limitation, the modular design allows individual components to be replaced or improved, providing a flexible foundation for future work, where the utility of these summaries could be validated through downstream tasks and human evaluation.

\bibliography{colm2026_conference}
\bibliographystyle{colm2026_conference}

\appendix

\section{Data Statistics}

\begin{table*}[h]
\centering
\small
\resizebox{\textwidth}{!}{
\begin{tabular}{|l|c|c|c|c|c|c|}
\hline
\textbf{Dataset} &
\textbf{Avg Utter. (duration)} &
\textbf{Avg Utter. (words)} &
\textbf{Avg Utter. (tokens)} &
\textbf{Avg Topic (duration)} &
\textbf{Avg Topic (words)} &
\textbf{Avg Topic (tokens)} \\
\hline
\textbf{AMI}     & 4.67 (GT) & 12.45 (GT) &16.51 (GT)  &171.02s (GT)& 532.91 (GT) & 567.66 (GT) \\
\hline
\textbf{IEMOCAP} & 4.37s (GT)& 11.62 (GT) & 15.72 (GT)&       59.03 (Pred) &  169.88 (Pred) & 229.54 (Pred) \\
\hline
\end{tabular}
}
\caption{Dataset statistics for AMI and IEMOCAP. GT denotes ground-truth annotations, while Pred indicates labels predicted by LLaMA 3.1 8B.}
\label{tab:dataset-distribution}
\end{table*}

We conducted statistical analyses covering average duration, word count, and token count at both the utterance and topic levels, see Table~\ref{tab:dataset-distribution}. All measurements for AMI are derived from ground truth. For IEMOCAP, utterance-level measures are based on ground truth, while topic-level segmentation is predicted by LLaMA 3.1 8B. These statistics characterize structurally distinct dialogue settings: AMI consists of long meeting topics, whereas IEMOCAP comprises shorter, more fragmented interactions.

\section{Topic Segmentation and Emotion Recognition Agent Performance}
\label{sec:appendix_agents}

Table~\ref{tab:topic_segmentation} reports the topic segmentation performance on the subset of the AMI corpus used in our experiments, evaluated using the standard metrics Pk and WindowDiff. Our implementation obtains Pk = 0.4126 and Wd = 0.4620. We note that topic segmentation in our framework is used only as a preprocessing step to structure long dialogues into topical segments, and the segmentation module can be replaced by alternative approaches without affecting the overall pipeline.

For a better understanding of the visual characteristics of AMI \ref{fig:ami_frames} and IEMOCAP \ref{fig:iemocap_frames}, we show example frames from both datasets. These examples illustrate that emotion recognition can be challenging due to variations in video quality, viewpoint, and facial visibility.

\begin{figure}[h]
    \centering
    \includegraphics[width=0.45\linewidth]{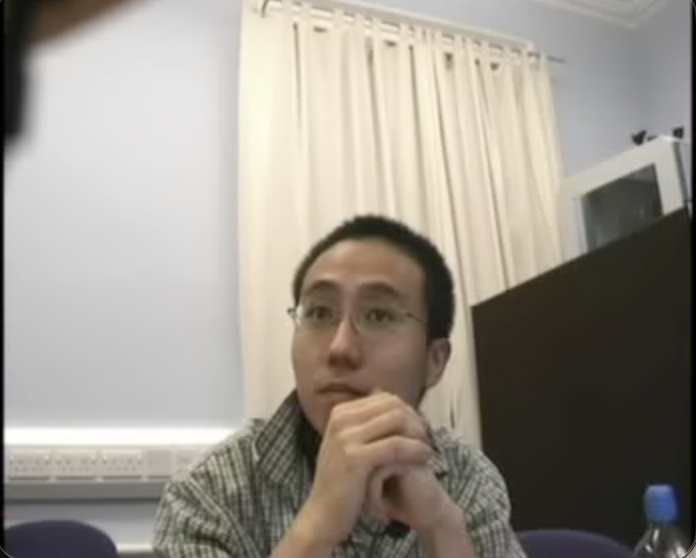}
    \hfill
    \includegraphics[width=0.45\linewidth]{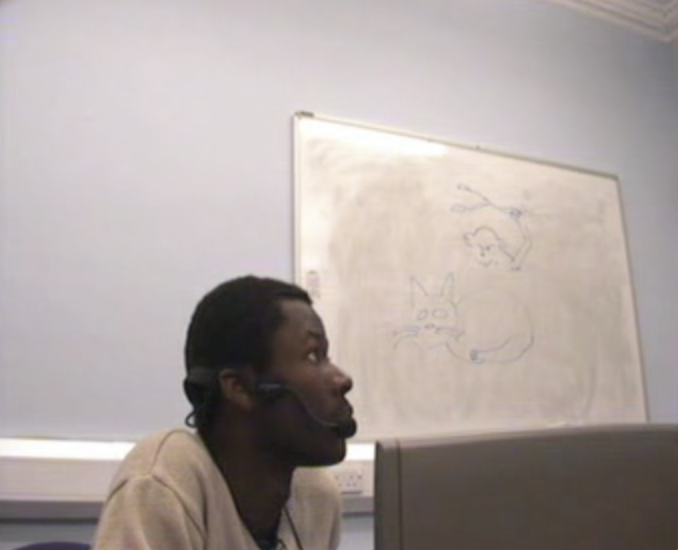}
    \caption{Example frames from AMI.}
    \label{fig:ami_frames}
\end{figure}

\begin{figure}[h]
    \centering
    \includegraphics[width=0.45\linewidth]{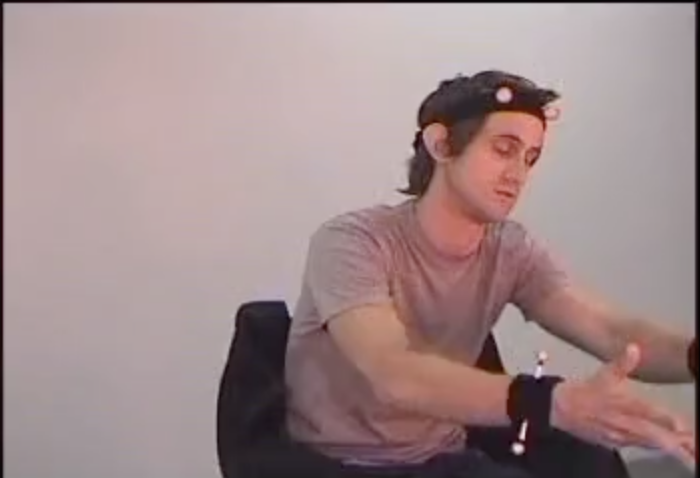}
    \hfill
    \includegraphics[width=0.45\linewidth]{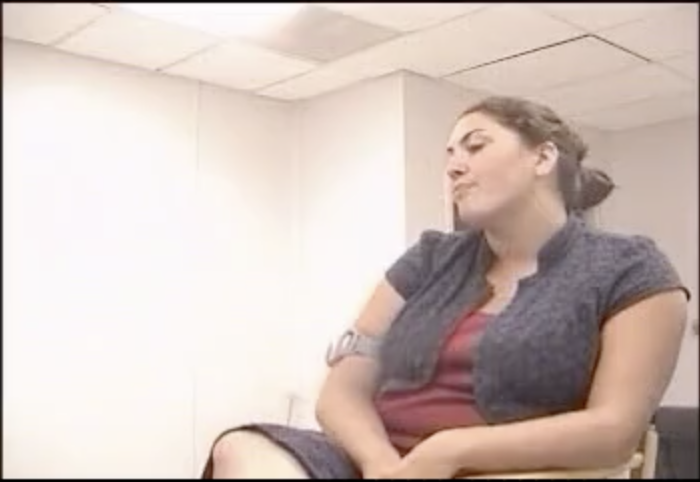}
    \caption{Example frames from IEMOCAP.}
    \label{fig:iemocap_frames}
\end{figure}

\begin{table}[h]
\centering
\small
\begin{tabular}{lcc}
\hline
\textbf{Method} & \textbf{AMI Pk} & \textbf{AMI Wd} \\
\hline
Random & 0.604 & 0.751 \\
Even & 0.513 & 0.543 \\
TextTiling & 0.391 & 0.410 \\
BiLSTM & 0.447 & 0.473 \\
BERT & \textbf{0.331} & \textbf{0.333} \\
S-BERT & 0.339 & 0.334 \\
\hline
Our (AMI subset) & 0.4126 & 0.4620 \\
\hline
\end{tabular}
\caption{Topic segmentation performance on the AMI dataset measured with Pk and WindowDiff (Wd). Lower values indicate better segmentation quality. Results of prior methods are reported from \citet{solbiati2021unsupervisedtopicsegmentationmeetings}.}
\label{tab:topic_segmentation}
\end{table}

\begin{figure}[t]
    \centering
    \includegraphics[width=0.9\linewidth]{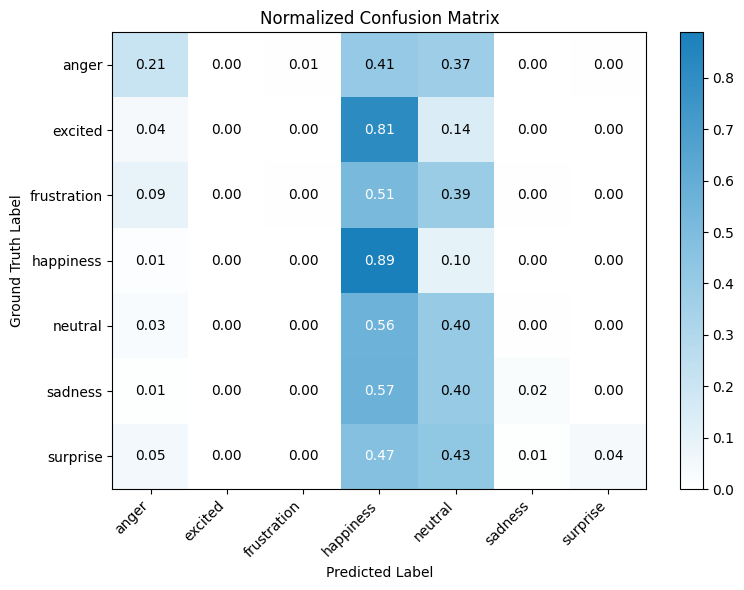}
    \caption{Confusion Matrix of Emotion-LLaMA on IEMOCAP}
    \label{fig:emotion_analysis}
\end{figure}

Figure \ref{fig:emotion_analysis} illustrates the confusion matrix of Emotion-LLaMA on the IEMOCAP dataset. The matrix reveals a strong tendency for the model to predict neutral and happiness across a wide range of ground-truth labels. This bias may be attributed to the higher frequency of these emotions in the AMI predictions.

\section{Examples for all the agents in our framework}
\label{app:agent_example}

\begin{topicseg}{Topic Segmentation Agent}\label{box:topicseg}
\footnotesize
\textbf{Input:} Dialogue transcript in JSON format \textit{D}

{\ttfamily
[ \\
\ \{"0": "utterance 1"\},\\
\ \{"1": "utterance 2"\},\\
\ ... \\
]
}

\textbf{Output:} Topic boundaries (utterance indices)\\
{\color{Accent}\textbf{Format:}}
$\{T_1:(0,12),\; T_2:(13,25),\dots\}$
\end{topicseg}

\begin{emo}{Emotion Recognition Agent}\label{box:emo}
\footnotesize
\textbf{Input:} Utterance-level, participant-specific text, audio, and video

\textbf{Output:} Emotion labels for each utterance\\
{\color{Accent}\textbf{Format:}}
$\{e_{U_1}:\text{happiness}\}$
\end{emo}

\begin{topicsum}{Topic Summarization Agent}
\footnotesize
\label{topicsum}
\textbf{Input:} Topic-level transcript with emotion labels assigned to each utterance
\footnotesize
\begin{verbatim}
T1: {
  "0": {
    "utterance": "It's time for me to go.",
    "speaker": "Speaker1",
    "emotion": "Sadness"
  },
  "2": {
    "utterance": "I can't believe that",
    "speaker": "Speaker2",
    "emotion": "Frustration"
  },
  ...
}
\end{verbatim}
\textbf{Output:} Topic summarization with emotion trajectory\\
{\color{Accent}\textbf{Example:}}
Speaker2 is trying to convince Speaker1 to let them go for a year due to government necessity, but Speaker1 refuses and expresses \textbf{sadness}. Speaker2 shows \textbf{frustration}, feeling overwhelmed by the situation and worried about their separation. Speaker1 attempts to comfort them while still struggling with their own \textbf{sadness}.
\end{topicsum}

\begin{speakersum}{Participant-specific Summarization Agent}
\footnotesize
\label{speakersum}
\textbf{Input:} Speaker-specific utterances with emotion labels
\footnotesize
\begin{verbatim}
P1: {
    "0": {
        "utterance": "It's time for me to go.",
        "emotion": "Sadness"
    },
    "1": {
        "utterance": "There's nothing I can do.  I mean, the government you know--",
        "emotion": "Sadness"
    },
  ...
}
\end{verbatim}
\textbf{Output:} Topic summarization with emotion trajectory\\
{\color{Accent}\textbf{Example:}}
The speaker was saying goodbye, feeling \textbf{sad} and worried about leaving their child with the father. They were concerned for the child's safety and wanted to ensure she had a normal childhood, despite having no control over the situation. With \textbf{sadness} and resignation, they accepted that there was nothing else they could do but leave."
\end{speakersum}

\begin{oversum}{Summary Aggregation Agent}
\footnotesize
\label{oversum}
\textbf{Input:} Summaries for each topic in order

{\footnotesize
\begin{verbatim}
{
  ST1: summary for topic 1,
  ...
  STn: summary for topic n,
  SP1: summary for participant 1,
  ...
  SPm: summary for participant m,
}
\end{verbatim}
}

\textbf{Output:} Dialogue summarization with emotion trajectory\\
{\color{Accent}\textbf{Example:}}
{
Speaker1 starts by expressing \textbf{sadness} due to the impending deployment, but eventually becomes more \textbf{excited} about the future and promises happiness for their baby. The conversation shifts from \textbf{sadness} to \textbf{excitement} and acceptance, with Speaker2 reassuring Speaker1 that they will return safely and stay in touch through letters.
}
\end{oversum}

\section{Emotion Trajectory Compression Algorithm}

\begin{algorithm}[h]
\caption{Emotion Trajectory Compression}
\label{alg:compress_emotion}
\begin{algorithmic}[1]
\Require Emotion trajectory \(\mathbf{e} = (e_1, \dots, e_T)\)
\Ensure Compressed trajectory \(\hat{\mathbf{e}}\)

\State Initialize \(\hat{\mathbf{e}} \gets [\,]\)
\If{\(T = 0\)}
    \State \Return \(\hat{\mathbf{e}}\)
\EndIf

\State Append \(e_1\) to \(\hat{\mathbf{e}}\)
\For{\(t = 2\) to \(T\)}
    \If{\(e_t \neq e_{t-1}\)}
        \State Append \(e_t\) to \(\hat{\mathbf{e}}\)
    \EndIf
\EndFor

\State \Return \(\hat{\mathbf{e}}\)
\end{algorithmic}
\end{algorithm}

\section{Full Pipeline results}
\label{tab:full_pipeline}
\begin{table*}[h
]
\centering
\small
\resizebox{\textwidth}{!}{
\begin{tabular}{l l ccccc ccccc}
\toprule

\multirow{3}{*}{\textbf{Model}} &
\multirow{3}{*}{\textbf{Setup}} &
\multicolumn{5}{c}{\textbf{AMI}} &
\multicolumn{5}{c}{\textbf{IEMOCAP}} \\

\cmidrule(lr){3-7} \cmidrule(lr){8-12}

& &
\multicolumn{1}{c}{\textbf{Content}} &
\multicolumn{4}{c}{\textbf{Emotion}} &
\multicolumn{1}{c}{\textbf{Content}} &
\multicolumn{4}{c}{\textbf{Emotion}} \\

\cmidrule(lr){3-3} \cmidrule(lr){4-7} \cmidrule(lr){8-8} \cmidrule(lr){9-12}

& &
\textbf{BL} & \textbf{LEV} & \textbf{NGR} & \textbf{JAC} & \textbf{COS} &
\textbf{BL} & \textbf{LEV} & \textbf{NGR} & \textbf{JAC} & \textbf{COS} \\

\midrule

\multirow{3}{*}{LLaMA 3.1 8B}
& {GT-assisted} & 0.0650 & 0.1060 & 0.5400 & 0.2500 & 0.2950 & 0.0810 & 0.2860 & 0.4680 & 0.2160 & 0.3130 \\
\cmidrule(lr){2-12}
& {Full pipeline}        & 0.0598 & 0.0938 & 0.5004 & 0.2234 & 0.2579 & 0.0798 & 0.2209 & 0.3729 & 0.1739 & 0.2268 \\
\cmidrule(lr){2-12}
& {$\Delta$}    & -0.0052 & -0.0122 & -0.0396 & -0.0266 & -0.0371 & -0.0012 & -0.0651 & -0.0951 & -0.0421 & -0.0862 \\

\bottomrule
\end{tabular}
}
\caption{
Comparison between the full pipeline and the ground-truth-assisted setting for LLaMA 3.1 8B model. $\Delta$ indicates the difference.
}
\label{tab:full_vs_gt_delta}
\end{table*}

In our experimental setup, we use ground-truth annotations when available to reduce confounding effects. We also evaluate the fully automated pipeline, which incorporates all agent components. Table \ref{tab:full_vs_gt_delta} shows that performance decreases under the full pipeline across both content and emotion trajectory metrics, likely due to inaccuracies of the Topic Segmentation and Emotion Recognition Agent in the Multimodal Understanding Layer.

\section{LLM Usage Disclosure}
In accordance with the COLM 2026 policy on LLM usage, we disclose that this work used LLMs for emotion trajectory evaluation generating code for plots.

\section{Reproducibility Statement}

We ensure reproducibility by providing a complete implementation of our dialogue summarization framework with emotion dynamics, including all components, data processing steps, and evaluation procedures. The repository is available at: \url{https://anonymous.4open.science/r/Dialogue_summarization_with_Emotion_Dynamics-649A/}.

\paragraph{Code Structure.}
The repository is organized as follows:
\begin{verbatim}
dynamic_summary/
    dataset/                  % raw and processed data
    preprocessing/            % data preprocessing scripts
    project/
        task_segmentation/    % topic segmentation module
        summarization_components/
            topic_summarization.py
            individual_summarization.py
            overall_summarization.py
            emotion_classification.py
            blanc.py
            metrics.py
            combined_eval.py
    config.py
    main.py
    requirements.txt
\end{verbatim}

Each module corresponds to a stage in the pipeline, including topic segmentation, emotion recognition, and summarization (topic-level, individual-level, and overall summarization). Emotion recognition is implemented using the Emotion-LLaMA model\footnote{\url{https://github.com/ZebangCheng/Emotion-LLaMA}}, and users should follow the installation instructions provided in the original repository.

All components can be executed independently, and the \texttt{main.py} script enables running the full pipeline end-to-end. Intermediate outputs are stored to facilitate inspection and verification of each stage. The project can be installed using the \texttt{requirements.txt} file. Users are required to update the configuration file (\texttt{config.py}) to match their local data paths.

\paragraph{Data Processing.}
We provide preprocessing scripts that transform raw dialogue datasets into the required input format for each module. \paragraph{Prompts and Model Configuration.} All prompts used for LLM-based components (e.g., segmentation, emotion extraction, and summarization) are included in the repository. All LLaMA-based models are executed with an extended context window of 32\,768 tokens. \paragraph{Evaluation.}
All evaluation metrics are implemented in the codebase, including BLANC and our proposed emotion trajectory-based metrics. The \texttt{combined\_eval.py} script provides a unified interface to compute all metrics and stores the results in a structured format (e.g., CSV or Excel) for analysis.

\paragraph{Reproducibility Considerations.}
While we fix configurations and provide complete inputs and prompts, minor variations may arise due to nondeterministic behavior in large language model inference or differences in hardware environments. We mitigate this by documenting all settings and providing deterministic preprocessing.

\end{document}